# YOLOatr : Deep Learning Based Automatic Target Detection and Localization in Thermal Infrared Imagery

Aon Safdar[1], Usman Akram[1], Waseem Anwar[2], Basit Malik[1], Mian Ibad Ali[3]

[1]*National University of Sciences and Technology, Islamabad, Pakistan*
[2]*Mälardalen University, Västerås, Sweden*
[3]*University of Galway, Ireland*

**Abstract**

Automatic Target Detection (ATD) and Recognition (ATR) from Thermal Infrared (TI) imagery in the defense and surveillance domain is a challenging computer vision (CV) task in comparison to the commercial autonomous vehicle perception domain. Limited datasets, peculiar domain-specific and TI modality-specific challenges i.e., limited hardware, scale invariance issues due to greater distances, deliberate occlusion by tactical vehicles, lower sensor resolution and resultant lack of structural information in targets, effects of weather, temperature, and time of day variations and varying target to clutter ratios all result in increased intra-class variability and higher inter-class similarity making accurate real-time ATR a challenging CV task. Resultantly, contemporary state-of-the-art (SOTA) deep learning architecture under-perform in the ATR domain. We propose a modified anchor-based single-stage detector called YOLOatr, based on a modified YOLOv5s, with optimum modifications to detection heads, feature-fusion in the neck, and a custom augmentation profile. We evaluate the performance of our proposed model on a comprehensive DSIAC MWIR dataset for real-time ATR over both correlated and decorrelated testing protocols. The results demonstrate that our proposed model achieves state-of-the-art ATR performance of up to 99.6%.

## 1 Introduction

Object detection, localization, classification, and tracking are the key components of modern computer-vision-based applications. Automatic Target Detection (ATD) encompasses the detection and localization of the objects in an image while Automatic Target Detection (ATR) involves further classification of the detected objects into relevant classes. Both terms (ATD and ATR) are usually associated with object detection/ recognition in surveillance and defense domains where robust target recognition and tracking are considered critical. Conventional imaging modalities in these domains include Infrared (IR) and synthetic aperture radar (SAR) with Thermal IR (TIR) being the predominant modality for ground-based tactical platforms with proven benefits [Zhao et al., 2022]. TIR includes shortwave infrared (SWIR 1-3 μm) and mediumwave infrared (MWIR, 3-5 μm) bands of the infrared wavelength where most captured radiations are emitted and not reflected from the targets [Berg et al., 2015]. TIR thus provides the ability to see targets both in the presence or absence of light. It is a frequently used imaging modality for ground based tactical/surveillance vehicles.

Despite the numerous advantages offered by TIR images, object detection/recognition using IR modality in tactical/surveillance applications suffers from peculiar domain challenges that affect reliability and robustness of an ATR system [Berg et al., 2015]. Ground-based tactical platforms generally operate in extreme weather and temperature conditions. This affects the image sensor output thus degrading heat signature differential between the target and the background [Liang et al., 2022]. The occlusion in defense scenario is deliberate as the target tries to camouflage or conceal itself and attempts to be occluded from view. Scale variations are high in case of ATR as the sensor-to-target distances and much greater. Low resolutions at far distances make discerning the defining target features difficult resulting in high inter-class similarities. Moreover, the environment is highly cluttered that makes detection/recognition a daunting task [Arif & Mahalanobis, 2020a]. Contextually, ATR has to be performed on a live video and not on still images. In videos, motion blur and occlusion are more vital than in still images. Moreover, fast inference speeds for real-time performance becomes critical. ATD/R has to be performed in varying practical scenarios with respect to movement of source and target. Consequently, there is high viewpoint, occlusion and scale variability in practical scenarios. Furthermore, IR data captured by the sensor varies significantly with metrological conditions, signal-to-clutter noise ratio (SCNR), time of day, sensor calibration and target viewpoint variations that increase the intra-class variabilities [Batchuluun et al., 2020]. For defence and surveillance applications, a significant problem is non-availability of comprehensive IR datasets that can be used for training and evaluation of ATR algorithms. The lack of suitable large-scale labeled IR dataset for military / surveillance objects is attributable to confidentiality. Owing to these reasons highlighted above, latest state of the art (SOTA) algorithms not only remain under-evaluated for TIR-based ATR, they also under-perform due to peculiarities specific to the IR modality and to the tactical domain. In this paper, we investigate the application of a state-of-the-art Deep Convolutional Neural Network (DCNN) based detector to address target detection, recognition and localization in thermal infrared video imagery for defence domain. Specifically, we explore different learning approaches, architectural changes and data augmentation profiles with a single-stage, single-frame object detector called You Only Look Once – Version 5 (YOLOv5) and propose the optimum learning approach, architectural changes and data augmentations to improve accuracy and robustness. We evaluate our approach and proposed model on the Defense Systems Information Analysis Center (DSIAC) benchmark ATR dataset.

This paper is organized as: Literature review is provided in Section 2. The YOLOatr development methodology is explained in Section 3. Result Evaluation forms part of Section 4. The discussion and limitations are given in Section 5. Finally, the article is concluded in Section 6.

## 2 Literature Review

The learning-based target detection/recognition solutions can be categorized into traditional machine learning-based methodologies and deep learning-based methodologies. The prominent and pertinent dataset used for vehicle detection algorithm development in the tactical domain is the DSIAC ATR dataset. Deep neural networks applied to this dataset can be divided into two categories based on the testing range: correlated and decorrelated datasets. Correlated datasets involve testing on the same range as training, while decorrelated datasets involve testing on higher ranges.

[Mahalanobis & McIntosh, 2019] compare the performance of Faster R-CNN network with a quadratic correlation filter (QCF) based algorithm using the DSIAC MWIR dataset. Millikan et al., 2018 proposes QCF filters in the first layer of a CNN for target classification but does not discuss ATR at longer ranges with high clutter. A different approach [d'Acremont et al., 2019] addresses the issue of inadequate datasets for training CNNs by using a compact and fully convolutional neural network with global average pooling, along with a simulated/synthetic dataset. This approach shows improved recognition performance and robustness to scale and viewpoint variations. Another study [Arif & Mahalanobis, 2020] employs a CNN-based auto-encoder to generate unseen views for the DSIAC dataset, achieving a test accuracy of 68% for a single vehicle class at a 1000m range. Similarly, an autoencoder and Siamese network are used [Arif & Mahalanobis, 2021] to generate realistic images from the DSIAC dataset. Another experiment [Chen et al., 2021] compares three variants of a single-stage detector with varying backbone structure for feature extraction, achieving high mean Average Precision (mAP) for ATR using four target classes.

For decorrelated datasets, various studies [McIntosh et al., 2020a/b; Jiban et al., 2021; Cuellar & Mahalanobis, 2021] argue that training and testing over the same ranges introduce potential biases, as clutter diversity, scale, and resolution issues affect detector performance. These representative studies split the DSIAC dataset for training at lower and testing at higher ranges. One study [McIntosh et al. 2020] proposes TCRNet, a custom network utilizing analytically derived eigen filters to maximize the Target-to-Clutter (TCR) metric. Another study [Jiban et al., 2021] improves upon TCRNet by processing target and clutter information in parallel channels before combining for TCR metric optimization. However, these studies focus on target detection (ATD) rather than target classification (ATR).

A few studies [Millikan et al., 2018 ; Liang et al., 2022] use multi-frame techniques based on background consistency and target sparsity hypotheses for moving target detection in the DSIAC datasets. However, these hypotheses are violated in real-time scenarios. In such cases, single-frame detection techniques are considered more practical

**Research Gap:** Based on literature review, the identified research gaps include:1) SOTA detectors largely remain under-evaluated and also under-perform in tactical setting due to peculiar IR and domain challenges. 2) There is a dearth of TIR based datasets for the ATR community. Moreover, annotating and preparing large video-based datasets (such as DSIAC dataset) requires tremendous amount of time and human effort and is required to be automated. 3) A robust single-frame, single-stage, anchor-based detector is required for high-accuracy, high-speed inference using limited hardware onboard tactical vehicles with generalization ability over decorrelated ranges, at multiple-combined distances and in varied illumination/SCNR (Day/Night) conditions. To the best of the author's knowledge there are no studies to date that have improved SOTA single stage detector, specifically YOLOv5, for ATR task on the DSIAC MWIR dataset at both correlated and decorrelated ranges with combined distances and time (both day and night). Our goal is to improve YOLOv5 for a robust solution to reliably detect small IR targets in high clutter at multiple ranges simultaneously in all weather, illumination, range and, viewpoint variations.

# 3 YOLOatr Development Methodology

Our YOLOatr development methodology (Figure 1) is divided into two parts. Firstly, we preprocess the DSIAC dataset for our experiments. Secondly, we follow an experimental approach and perform an ablation study over various model variants of the base model (YOLOv5s) to find the optimum modifications in terms of training methodology, architecture and data augmentation. Finally, we use the optimum model (YOLOatr) and evaluate its performance and generalization ability. We finally perform a comparative analysis of results achieved by YOLOatr with other SOTA detectors used in various studies.

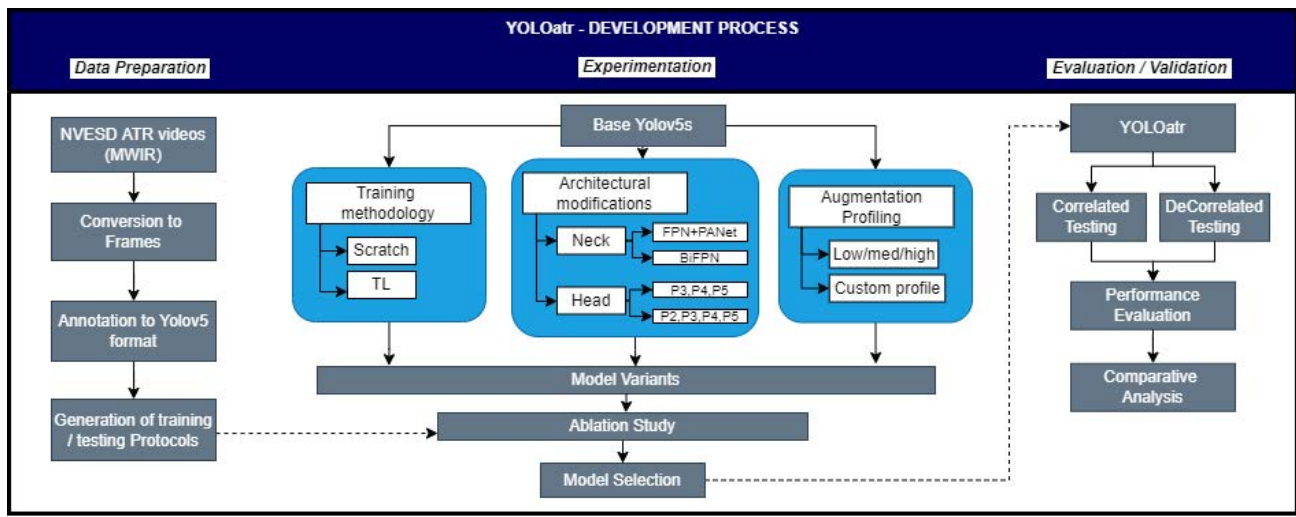

**Figure 1 : Methodology for development of YOLOatr**

## 3.1 Dataset and Partitioning Protocols

**DSIAC MWIR Dataset:** We have utilized the US Army Night Vision and Electronic Sensors Directorate's (NVESD) ATR Algorithm Development Image Database which is the largest and most comprehensive publicly available TIR dataset for ground-based tactical platforms. The dataset was released by the U.S Department of Defence (DoD) for ATR algorithm development and contains visible and mid-wave infrared (MWIR) thermal images of tactical vehicles over long ranges in the desert environment. It contains (Table 1) images of 13 different tactical and civilian vehicles moved from 1000m to 5000m with increments of 500m. Data is collected for both day and night situations. 72 aspect angles for each vehicle type are obtained by driving in a circle of 100m at a speed of 10mph. Minute-long videos @ 30fps for each vehicle during both day and nighttime were converted into frames and annotated for ground truth. Four different vehicles (T72, BRDM2, Pickup, SUV) were used for our experiments (Figure 2). A total of 3600 images were used for each vehicle type with 70% data for training, 20% for validation, and 10% for testing.

|  | MWIR | Visible |
|---|---|---|
| **Source Camera** | "cegr" | "i1co" |
| **Size** | 207 GB | 106 GB |
| **Aspect Angles** | 72 - obtained by driving in circle of 100m @ 10mph | |
| **Distances** | 1000m to 5000m @ increments of 500m for both day and night | |
| **Files** | 186 Files in ARF format viewable using ImageJ software via a plugin, ground truth in ATG format | |
| **Length** | 1 min videos @ 30fps (1800 frames per video) | |
| **Targets** | 10 Vehicles (2 civilian and 8 tactical vehicles) 2 Human (slow- and fast-moving pedestrians) | |
| **Samples (one target)** | 1800 frames x 18 videos = 32400 frames (Vehicle) 1800 frames x 12 videos = 21600 frames (Human) | |
| **Total Dataset size** | (32400 x 10) + (21600 x 2) = 367,200 approx. | |

**Table 1 : The DSIAC dataset**

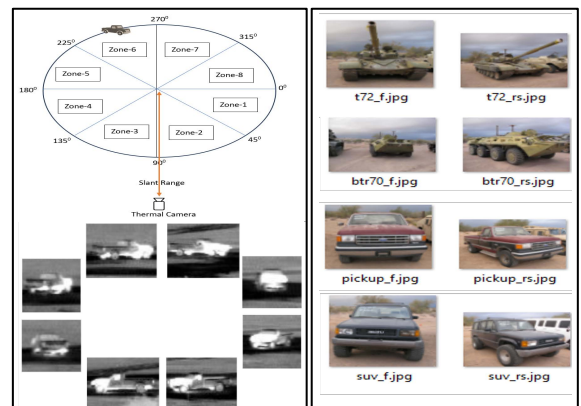

**Figure 2 : Image Acquisition (Left) and Selected Targets (Right)**

**Training and Testing Protocols:** To ascertain the performance and generalization ability of our model, we use two testing protocols. The 'correlated dataset partitioning' protocol (T1) where training and testing images are from the same range and the 'decorrelated dataset partitioning' protocol (T2) where training is done at lower ranges and testing at higher ranges. The details of partitioning protocols are given in Table 2.

### 3.2 YOLOv5s Architectural Modifications

ATR for TIR requires high accuracy and real-time inference speed. This dictates the choice of selecting an appropriate Object Detector that strikes a balance between both speed and accuracy. YOLO (You Only Look Once) is a cutting-edge single-stage, single frame object detector that can achieve both goals (i.e., speed and accuracy). The YOLO family has many variants (e.g., YOLOv1, YOLOv2, YOLOv5, YOLOX, YOLOR etc.). For our base model we selected YOLOv5s, which is the small variant of the Yolov5 series [Jocher, 2022]. We experimented with architectural modifications to the original YOLOv5s neck and head. As our dataset contains great concentration of small objects with only few pixels on target and because YOLOv5 is known to struggle with small object detection, we introduced an extra small detection head (P2) in order improve small object detection performance. Moreover, we replaced the original PANet neck with BiFPN (Weighted Bi-directional Feature Pyramid) Network neck to promote better feature fusion and propagation. Our modified YOLOv5s architecture is shown in Figure 3.

| Dataset | Training (Day + Night) | | | Testing (Day + Night) | | | | | |
|---|---|---|---|---|---|---|---|---|---|
| | | | | Correlated (T1) | | | Decorrelated (T2) | | |
| | Range in Km | Images | | Range in Km | Images | | Range in Km | Images | |
| | | One Target | Four Targets | | One Target | Four Target | | One Target | Four Targets |
| DS1 | 1.0, 1.5, 2.0, 2.5 | 10080 | 8064 | 1.0, 1.5, 2.0, 2.5 | 2880 | 2304 | 3.0 | 3600 | 2880 |
| DS2 | 3.0, 3.5, 4.0, 4.5 | 10080 | 8064 | 3.0, 3.5, 4.0, 4.5 | 2880 | 2304 | 5.0 | 3600 | 2880 |

**Table 2 : Dataset partitioning protocols**

### 3.3 Learning Methodology

We experimented with two learning methodologies two learning methodologies i.e, learning from scratch and transfer learning. We tested both approaches initially with baseline YOLOv5s. For learning from scratch, we trained the model using 100 epochs with randomly initialized weights. We used the training images as per established protocol for training from scratch. For transfer learning we used pre-trained weights (Imagenet) and trained with the backbone layers frozen for 30 epochs, followed by 10 epochs for fine-tuning in which all layers were unfrozen. The results with training from scratch were slightly better than from transfer learning. This may be attributed to the fact that the weights learned from RGB images do not comply with features present in the TIR objects. Using weights/knowledge from RGB domain for the TIR domain did not offer any advantage except for reduced training time/ epochs.

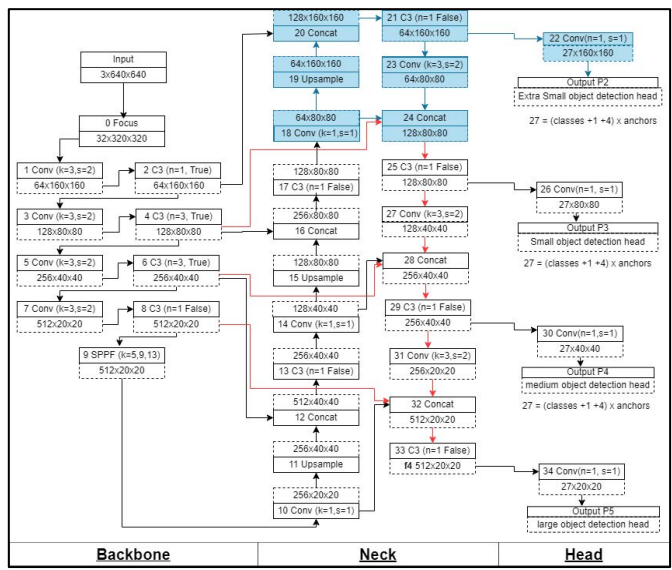

**Figure 3 : Architectural modifications to YOLOv5**

### 3.4 Augmentation Profiling

YOLOv5s uses a rich set of data augmentation techniques at training time to improve model accuracy by diversifying input images. Considering the nature of available images and ATR task, we created a custom augmentation profile (CAP) for model training as highlighted in Table.3.

| Augmentation type | Custom value |
|---|---|
| *fl_gamma* – Focal Loss Gamma | 0.3 |
| *hsv_h* - Hue | 0.015 |
| *hsv_s* - Saturation | 0.7 |
| *hsv_v* - Value | 0.4 |
| *degrees* – Rotation (+/- degree) | 3 |
| *translate* – Translation (+/- fraction) | 0.1 |
| *Scale* – Image Scale (+/- gain) | 0.3 |
| *shear* – Image Shear (+/- deg) | 0.0 |
| *perspective* (+/- fraction) | 0.0005 |
| *flipud* - flip up/down | 0.1 |
| *fliplr* - flip left/right | 0.5 |
| *mosaic* | 0.1 |
| *mixup* | 0.4 |
| *copy_paste* | 0.5 |

**Table 3 : Custom Augmentation Profile**

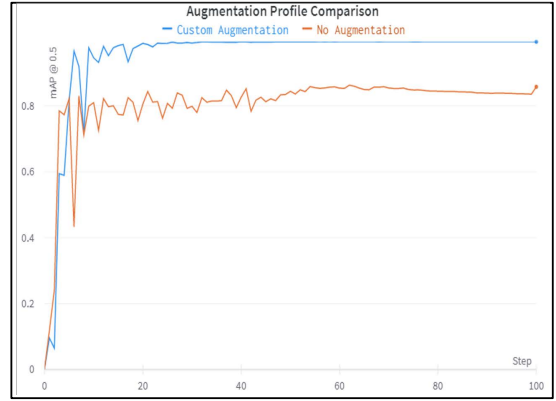

**Figure 4 : Model Training Accuracy Comparison of Default vs CAP**

Specifically, the brightness and contrast changes were kept as they are in line with varied illumination conditions caused by weather and time of day/night in practical scenarios. Similarly, the translation, rotation, scaling, flipping, and perspective changes were kept at medium level as ample variations in translation and aspect angles are available in the original dataset. The mosaic/ tiling data augmentations helps model with small object detection. It was kept low as the targets in original images at long ranges occupy only a few pixels and reducing their size further would be counterproductive. Similarly, as the structural information in low resolution original images at long ranges is already very low, the shear augmentation was turned off. Mix-up and copy-paste augmentations cross-stitch samples to increase the richness/ diversity of data and were kept high for that reason. YOLOv5s also provides albumentations integration for added augmentation of blurring etc. However, albumentations being additive to data augmentations were turned off so as not to overwhelm the model training process. Table 3 details the changes incorporated through our custom augmentation profile. Figure 4 shows how the CAP improves model training and validation accuracy in comparison with no/low data augmentation.

## 4 Evaluation and Validation

As the YOLOv5s model has not been previously used with the DSIAC dataset, we first performed experiments with the default YOLOv5s model to ascertain baseline performance of this cutting-edge detector. We then we incorporated our proposed architectural changes to neck and head, along with our custom data augmentation to train YOLOatr with our selected learning methodology and compared performance with the baseline model.

### 4.1 Experimental Setup and Evaluation Metrics

All experiments were performed on Google Colab pro platform using Nvidia Tesla P100-PCIE (16 GB) GPU and 32 GB High-speed RAM. The input image size was kept to 640x640. SGD optimizer was used with default learning rate and momentum instead of Adam because SGD generalizes better while Adam converges faster. The batch size was kept 32 and models were trained for 100 epochs. The commonly used evaluation metrics of precision (1), recall (2) and mean Average Precision (mAP) (3) at an Intersection over Union (IoU) threshold of 0.5 was used for ascertaining performance. The tactical/ATR domain equivalent of Precision is called 'Probability of target declaration (Pdc)', while Recall is analogous to 'Probability of detection (Pd)'. However, we use the commonly used terms of precision and recall in all our results.

$$Precision = \frac{TP+TN}{TP+TN+FP+FN} = \frac{TP}{All\ Detections} \qquad (1)$$

$$Recall = \frac{TP}{TP+FN} = \frac{TP}{All\ Ground\ Truths} \qquad (2)$$

$$mAP = \frac{1}{|Classes|} \sum_{i=1}^{|Classes|} AP^i \qquad (3)$$

## 4.3 Experimental Results

We ascertained the performance with both the original YOLOv5s and proposed YOLOatr on DS1 subset of the DSIAC ATR dataset. Both models were trained from scratch for 100 epochs with SGD optimizer. The results are shown in Table 4 (Left) showing that YOLOv5s performed commendably for the correlated test range (T1) for all target types achieving a mAP score of 99.4% while YOLOatr achieved a mAP of 99.6% with a 0.02% performance gain over YOLOv5s. The results establish the dominance of single-frame, single-stage, anchor-based DCNN in achieving high accuracy with a lean structure.

| Model | Training Range | Time | Testing Results (mAP) | | Parameters | Inference time (ms) | GFLOPs |
|---|---|---|---|---|---|---|---|
| | | | T1 (1.0-2.5 km) | T2 (3.0 km) | | | |
| YOLO5s | 1.0-2.5 | Day and Night | 0.994 | 0.263 | 7020913 | 4.0 | 16.2 |
| YOLOatr | | | 0.996 | 0.377 | 7086449 | 4.5 | 16.4 |
| Performance Gain | | | +0.02% | +11.4% | | | |

| Targets | Testing Protocol T1 | | | Testing Protocol T2 | | |
|---|---|---|---|---|---|---|
| | Precision | Recall | mAP@0.5 | Precision | Recall | mAP@0.5 |
| T72 Tank | 0.996 | 0.995 | 0.997 | 0.699 | 0.618 | 0.622 |
| BTR70 | 0.999 | 1.00 | 0.996 | 0.169 | 0.132 | 0.214 |
| SUV | 0.996 | 0.993 | 0.994 | 0.562 | 0.373 | 0.393 |
| Pickup | 1.00 | 0.999 | 0.995 | 0.278 | 0.435 | 0.163 |
| All | 0.996 | 0.997 | **0.996** | 0.512 | 0.44 | **0.377** |

**Table 4 : Performance of Yolov5 and YOLOatr (Left), YOLOatr Target-wise performance comparison (Right)**

However, the model accuracy of YOLOv5s declined significantly when tested over the decorrelated range (T2) achieving a mAP of 27.1%. The results highlight that although the YOLOv5s performs well over correlated ranges, it struggles to generalize over the decorrelated ranges. The probable reasons maybe that as range increases the scale becomes smaller and the model struggles to find small targets. Moreover, the clutter changes with decorrelated range and resolution, structure of the target also degrades at higher range. Here, YOLOatr outperformed YOLOv5s with a performance gain of 11.4% achieving mAP of 37.7%. Assessing the target-wise performance of YOLOatr in Table 4 (Right), the most affected target was Pickup truck with a mAP of 16.3% while Tank T-72 achieved highest accuracy at T2 with 62.2% score. This may be attributed to size and thermal signature which is more for the tactical vehicle as compared to the commercial vehicles.

Overall, YOLOatr shows a performance gain of 0.02% for ATR at decorrelated ranges and 11.4% for decorrelated ranges against YOLOv5s. The results achieved by YOLOatr are SOTA to date with near-perfect accuracy of 99.6% for small objects over long ranges in highly cluttered IR images. Performance of YOLOatr is highlighted in the Precision and Recall curves, confusion matrix and mAP (PR Curve) for unseen data in Figure 5 (Left). The results highlight that YOLOatr achieves SOTA performance for the ATR task.

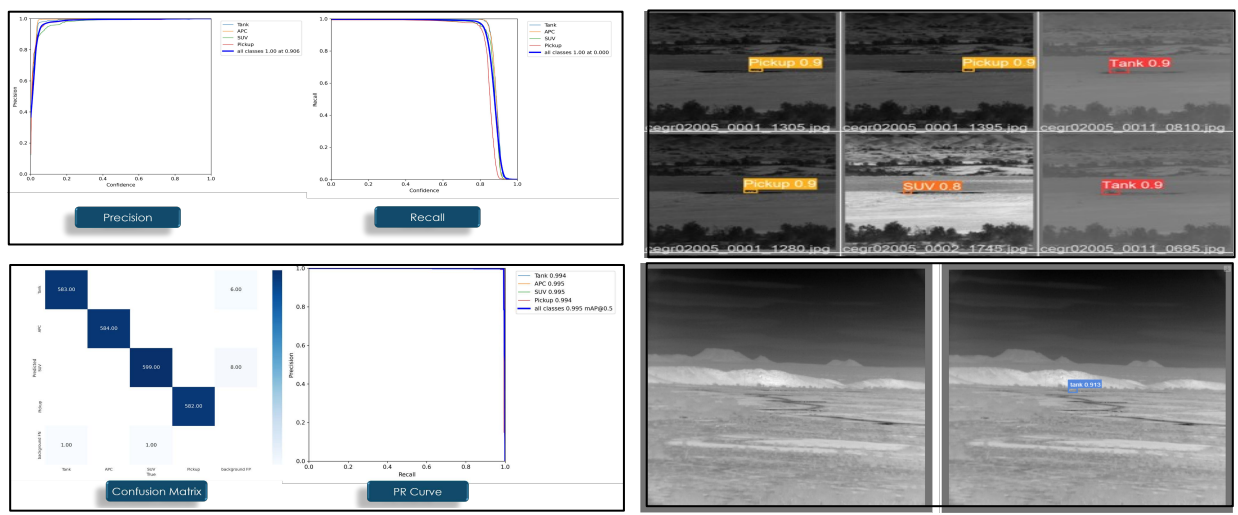

**Figure 5 : YOLOatr performance on Test data (Left), Detection of different targets (Top Right), Target detection at 5000m (Bottom Right, Detection at 5000m range (Bottom Right)**

Visual inference results for different targets are shown in Figure 5 (Right). YOLOatr detects target in highly cluttered background with a high confidence which is a prohibitive task for human vision. Considering the inherent challenges of IR modality and tactical domain, the performance of YOLOatr is phenomenal.

# 5 Discussion and Limitations

In this paper we have attempted to solve the problem of ATR in TIR for the tactical and surveillance domain. Specifically, we have developed, trained and tested YOLOatr, a lean, accurate, fast, and robust model for recognizing targets in challenging TIR images in real time. We have demonstrated that our model is able to detect robust classification features achieving up to 99.6% accuracy at distances up to 5000m. This is a prohibitive task for human vision but YOLOatr can both reliably detect and recognize different targets. The reliability of our ATR algorithm to detect and recognize each individual vehicle from as far as 5,000 meters is SOTA. Moreover, as compared to YOLOv5, YOLOatr improves the generalization ability over decorrelated ranges.

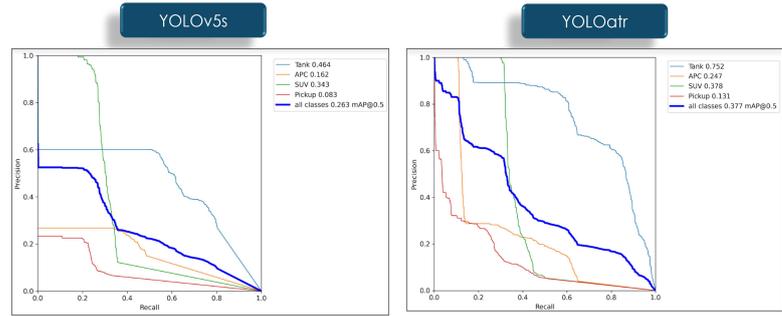

**Figure 6 : Performance improvement of YOLOatr over YOLOv5s**

An analysis of YOLOv5s performance over the difficult problem of decorrelated testing and improvements shown by YOLOatr is shown in Figure 6. It can be seen that YOLOatr improves results over decorrelated ranges and generalizes better than original YOLOv5s. A few other studies attempt to solve the problem with different techniques. However, mostly, the studies target ATD which only includes target detection and not recognition. Moreover, some studies only use nighttime data for result evaluation while most studies only use testing at correlated ranges only. A comparative analysis of YOLOatr with leading studies is drawn in Table 6. It can be seen that significant improvement wis achieved in terms of accuracy and speed.

| Ser | Publication | ATR/D | Classes | D/N | Range (km) | | Performance | | Performance YOLOatr | |
|---|---|---|---|---|---|---|---|---|---|---|
| | | | | | Tr | Te | mAP | Speed (fps) | mAP | Speed |
| 1. | Chen et al., 2021 | ATR | 4 | N | 4.0 | | 98.0% | 92.6-99.3* | 99.4% | |
| | | | | N | 5.0 | | 99.5% | | 99.5% | |
| | | | | D | 5.0 | | 85.5% | | 88.3% | |
| | | | | N | 4.0-5.0 | | 99% | | 99.4% | |
| 2. | Ceullar et al., 2021 | Yolov3 (SF) | ATD | 1 | D+N | 4.0-4.5 | | 95.87% | - | 99.6% | 110@ |
| | | Yolov3 (MF) | | | | 4.0-4.5 | 5.0 | 3.52% | - | 73.3% | |
| | | Mask-RCNN (SF) | | | | 4.0-4.5 | | 20.68% | - | 73.3% | |
| | | | | | | 4.0-4.5 | | 93.67% | - | 99.6% | |
| | | | | | | 4.0-4.5 | 5.0 | 1.34% | - | 73.3% | |
| | | MTINet | | | | 4.0-4.5 | 5.0 | 95% | - | 78% | |
| 3. | Maliha et al., 2021 | ATR | 10 | | 1.0-2.0 | | 88% | - | 98% | |
| 4. | Acremont et al., 2019 | ATR | 10 | | 1.0 | | 70% | - | 99% | |
| 5. | McIntosh et al., 2020 | ATR | 10 | D+N | 1.0-2.0 | 2.5-3.5 | 83% | - | 85.5% | |
| * Using NVIDIA Quadro RTX5000 @480x480 image size | | | | | | | | | @Using NVIDIA PC100 | |
| Other models using Geforce GTX Titan X @ 480x480 image size: <br> • Yolov2 - 59 fps <br> • Faster-RCNN with VGG16 – 7 fps <br> • Faster-RCNN with Resnet – 5 fps | | | | | | | | | Underlined: denotes Accuracy | |

**Table 5 : Performance comparison of YOLOatr with SOTA models on DSIAC dataset**

YOLOatr proves to be a robust single-stage detector that performs exceedingly well for both ATR and ATD cases to reliably detect targets in TIR images. It is a lean architecture with very fast inference speed of 110 fps making it a viable detector with high accuracy and less computational requirements. YOLOatr has few limitations. Firstly, optimum hyperparameter selection has not been explored via genetic algorithm available with yolov5. Secondly, comparative analysis of model accuracy with False Alarm Rate (FAR) has not been performed as done by few other studies. Finally, modifications to the head structure are not explored that may improve feature extraction. Performance over recently proposed variants of YOLO (Yolov7 Yolov8, yolo NAS etc.) is under experimentation and will form part of our research in future.

# 6 Conclusion

In this paper, we aim to achieve robust Automatic Target Detection / Recognition of small IR objects in cluttered background. We strive to do so over long distances, in cluttered background, and in varying illumination conditions. We focus on improving detection and classification accuracy with fast inference speed using a leaner architecture compatible with limited computational power available on-board ground-based tactical and surveillance vehicles. We propose YOLOatr with optimum training approach, augmentation profile and structural modifications to Yolov5s and tailored for the ATR case in challenging tactical / surveillance TIR domain. We demonstrate improvement in detection accuracy while preserving implementation potential. Additionally, we demonstrate better generalization ability for improved robustness over different range, viewpoint, scale, clutter and illumination variations.

# References

[Zhao et al., 2022] Zhao, M., Li, W., Li, L., Hu, J., Ma, P., & Tao, R. (2022). Single-frame infrared small-target detection: A survey. IEEE Geoscience and Remote Sensing Magazine.

[Berg et al., 2015] Berg, A., Ahlberg, J., & Felsberg, M. (2015, August). A thermal object tracking benchmark. In 2015 12th IEEE International Conference on Advanced Video and Signal Based Surveillance (AVSS) (pp. 1-6).

[Liang et al., 2022] Liang, X., Liu, L., Luo, M., Yan, Z., & Xin, Y. (2022). Robust infrared small target detection using Hough line suppression and rank-hierarchy in complex backgrounds. Infrared Physics & Technology, 120, 103893.

[Arif & Mahalanobis, 2020] Arif, M., & Mahalanobis, A. (2020). View prediction using manifold learning in non-linear feature subspace. In MIPPR 2019: Pattern Recognition and Computer Vision (Vol. 11430, pp. 316–323). SPIE.

[Batchuluun et al., 2020] Batchuluun, G., Kang, J. K., Nguyen, D. T., Pham, T. D., Arsalan, M., & Park, K. R. (2020). Deep learning-based thermal image reconstruction and object detection. IEEE Access, 9, 5951-5971.

[Mahalanobis & McIntosh, 2019] Mahalanobis, A., & McIntosh, B. (2019). A comparison of target detection algorithms using DSIAC ATR algorithm development data set (Vol. 1098808, No. May 2019, p. 3). doi: 10.1117/12.2517423.

[Millikan et al., 2018] url: https://dsiac.org/databases/ Accessed (Online) Jun 2023.

[Millikan et al., 2018] Millikan, B., Foroosh, H., & Sun, Q. (2018). Deep convolutional neural networks with integrated quadratic correlation filters for automatic target recognition. In Proceedings of the IEEE Conference on Computer Vision and Pattern Recognition Workshops (pp. 1222-1229).

[d'Acremont et al., 2019] d'Acremont, A., Fablet, R., Baussard, A., & Quin, G. (2019). CNN-based target recognition and identification for infrared imaging in defense systems. Sensors, 19(9), 2040.

[Arif & Mahalanobis, 2020] Arif, M., & Mahalanobis, A. (2020). Multiple view generation and classification of mid-wave infrared images using deep learning. arXiv preprint arXiv:2008.07714.

[Arif & Mahalanobis, 2021] Arif, M., & Mahalanobis, A. (2021). Infrared target recognition using realistic training images generated by modifying latent features of an encoder-decoder network. IEEE Transactions on Aerospace and Electronic Systems, 57(6), 4448-4456.

[Chen et al., 2021] Chen, H. W., Gross, N., Kapadia, R., Cheah, J., & Gharbieh, M. (2021, March). Advanced Automatic Target Recognition (ATR) with Infrared (IR) Sensors. In 2021 IEEE Aerospace Conference (50100) (pp. 1-13). IEEE.

[McIntosh et al., 2020a] McIntosh, B., Venkataramanan, S., & Mahalanobis, A. (2020). Infrared target detection in cluttered environments by maximization of a target to clutter ratio (TCR) metric using a convolutional neural network. IEEE Transactions on Aerospace and Electronic Systems, 57(1), 485-496.

[McIntosh et al., 2020b] McIntosh, B., Venkataramanan, S., & Mahalanobis, A. (2020, October). Target Detection in Cluttered Environments Using Infra-Red Images. In 2020 IEEE International Conference on Image Processing (ICIP) (pp. 2026-2030). IEEE.

[Jiban et al., 2021] Jiban, M. J. H., Hassan, S., & Mahalanobis, A. (2021, September). Two-Stream Boosted TCRNet for Range-Tolerant Infra-Red Target Detection. In 2021 IEEE International Conference on Image Processing (ICIP) (pp. 1049-1053). IEEE.

[Cuellar & Mahalanobis, 2021] Cuellar, A., & Mahalanobis, A. (2021, September). Detection of Small Moving Ground Vehicles in Cluttered Terrain Using Infrared Video Imagery. In 2021 IEEE International Conference on Image Processing (ICIP) (pp. 1099-1103). IEEE.

[Baili, 2020] Baili, N. (2020). Automatic target recognition with convolutional neural networks.

[Jocher, 2022] Jocher, G. (2022). ultralytics/yolov5. Retrieved from https://github.com/ultralytics/yolov5